%% file: IEEEabrv.tex
\documentclass[lettersize,journal]{IEEEtran}
\usepackage{amsmath,amsfonts}
\usepackage{algorithmic}
\usepackage{algorithm}
\usepackage{array}
\usepackage[caption=false,font=normalsize,labelfont=sf,textfont=sf]{subfig}
\usepackage{textcomp}
\usepackage{stfloats}
\usepackage{url}
\usepackage{verbatim}
\usepackage{graphicx}
\usepackage{cite}
\usepackage{algorithm}
\usepackage{algorithmic}

\usepackage{amsmath}
\usepackage{amsfonts}
\usepackage{bm}

\usepackage{booktabs}
\usepackage{threeparttable}

\usepackage[switch]{lineno}
\usepackage{multirow}
\usepackage[table,xcdraw]{xcolor}
\usepackage{lineno}

\usepackage{color}
\usepackage{bbding}
\begin{document}

\title{Small Object Tracking in LiDAR Point Cloud: Learning the Target-awareness Prototype and Fine-grained Search Region}

\author{Shengjing~Tian, 
        Yinan~Han*, 
        Xiuping~Liu*, 
        Xiantong~Zhao
\thanks{Shengjing~Tian is with the School of Economics and Management, China University of Mining and Technology, Xuzhou, China (e-mail:tye.dut@gmail.com).}
\thanks{Yinan~Han is with the School of Mathematical Sciences, Dalian University of Technology, Dalian, China (e-mail: spolico.hyn@gmail.com, corresponding author).}
\thanks{Xiuping Liu and Xiantong~Zhao are with Dalian University of Technology (e-mail: xpliu@dlut.edu.cn,  934613764@qq.com), Xiuping Liu is also corresponding author.}
}

\markboth{Journal of \LaTeX\ Class Files,~Vol.~14, No.~8, August~2021}%
{Shell \MakeLowercase{\textit{et al.}}: A Sample Article Using IEEEtran.cls for IEEE Journals}


\maketitle

\input{sections/abstract}
\input{sections/introduction}
\input{sections/relatedworks}
\input{sections/methods}

\input{sections/experiments}
\input{sections/limitations}
\input{sections/conclusion}

\newpage
\bibliographystyle{IEEEtran}
\bibliography{IEEEabrv}

\end{document}

%% file: sections/abstract.tex
\begin{abstract}
Single Object Tracking in LiDAR point cloud is one of the most essential parts of environmental perception, in which small objects are inevitable in real-world scenarios and will bring a significant barrier to the accurate location. However, the existing methods concentrate more on exploring universal architectures for common categories and overlook the challenges that small objects have long been thorny due to the relative deficiency of foreground points and a low tolerance for disturbances. To this end, we propose a Siamese network-based method for small object tracking in the LiDAR point cloud, which is composed of the target-awareness prototype mining (TAPM) module and the regional grid subdivision (RGS) module. The TAPM module adopts the reconstruction mechanism of the masked decoder to learn the prototype in the feature space, aiming to highlight the presence of foreground points that will facilitate the subsequent location of small objects. Through the above prototype is capable of accentuating the small object of interest, the positioning deviation in feature maps still leads to high tracking errors. To alleviate this issue, the RGS module is proposed to recover the fine-grained features of the search region based on ViT and pixel shuffle layers. In addition, apart from the normal settings, we elaborately design a scaling experiment to evaluate the robustness of the different trackers on small objects. Extensive experiments on KITTI and nuScenes demonstrate that our method can effectively improve the tracking performance of small targets without affecting normal-sized objects. 
\end{abstract}

%% file: sections/introduction.tex
\section{Introduction}
\label{sec:introduction}

Single object tracking is one of the most classical tasks in computer vision, which can assist intelligent agents in understanding the environment to achieve more sophisticated goals. With the development and popularization of Light Detection And Ranging (LiDAR) technology, the visual object tracking community has also sparked a wave of 3D single object tracking with point cloud, which has been attracting more and more attention in multiple domains, such as autonomous vehicles and mobile robotics \cite{Comport2004RobustMT, Luo2018FastAF, Machida2012HumanMT}. 

According to the generation of proposals, existing 3D single object tracking networks can be divided into point-based and bev-based methods. The former directly utilize seed points sampled from the search region to regress bounding boxes \cite{Qi2020P2BPN}. The latter project points to a bird's eye view map and then predict 3D bounding boxes resorting to a 2D detection head, whose typical representatives include V2B and STNet \cite{Hui20213DSV, Hui20223DST}. Although bev-based methods inevitably suffer from information loss caused by projection, their facilitation during 2D methods migration and multimodal fusion still make it full of exploratory value. In this field, however, all of the existing methods focus on normal size objects but leave the small objects to be untapped. 

\begin{figure}
\begin{center}
\includegraphics[scale=0.5]{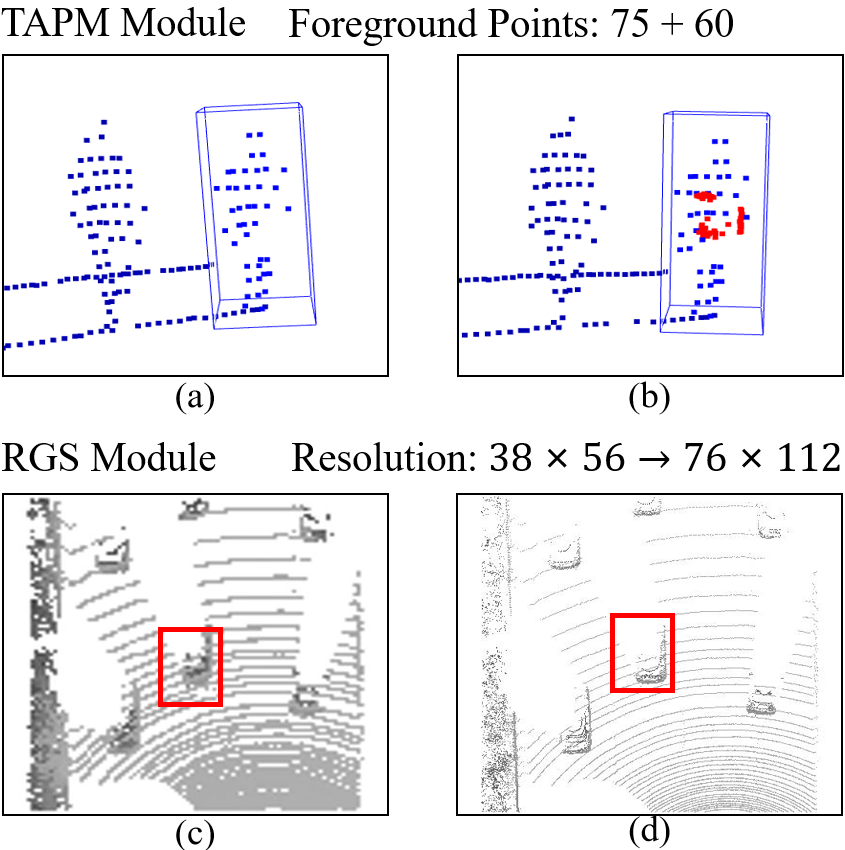}
\end{center}
\caption{The function of the proposed module. The target-awareness prototype mining module successfully recovers 60 foreground points, as (a) and (b) shown. The regional grid subdivision module upsamples the bird's eye view map with a resolution of $38 \times 56$ to $76 \times 112$ one, as (c) and (d) shown.}
\label{fig:contribution} 
\end{figure}

In order to shore up this weak link, this work shifts our efforts towards small objects which are common and inevitable in real-world scenarios. Generally, it is undoubted that pedestrians are categorized as typically small objects in the common datasets, such as KITTI \cite{Geiger2012AreWR} and nuScenes \cite{Caesar2019nuScenesAM}. In practice, we need to locate the specified object in $11.2m \times 7.2m$ search regions, where a pedestrian occupies no more than $1m \times 0.5m$ in contrast to a vehicle that usually covers a large area of over $4m \times 2m$. If a voxel size of 0.3m is used to encode the search area, then in a $ 24 \times 38$ pseudo image, such a small pedestrian will be only equivalently transformed into an object close to $3 \times 2$ pixels, which strictly conforms to the definition of small objects in the literature \cite{Chen2016RCNNFS}.

\begin{figure}
\begin{center}
\includegraphics[scale=0.33]{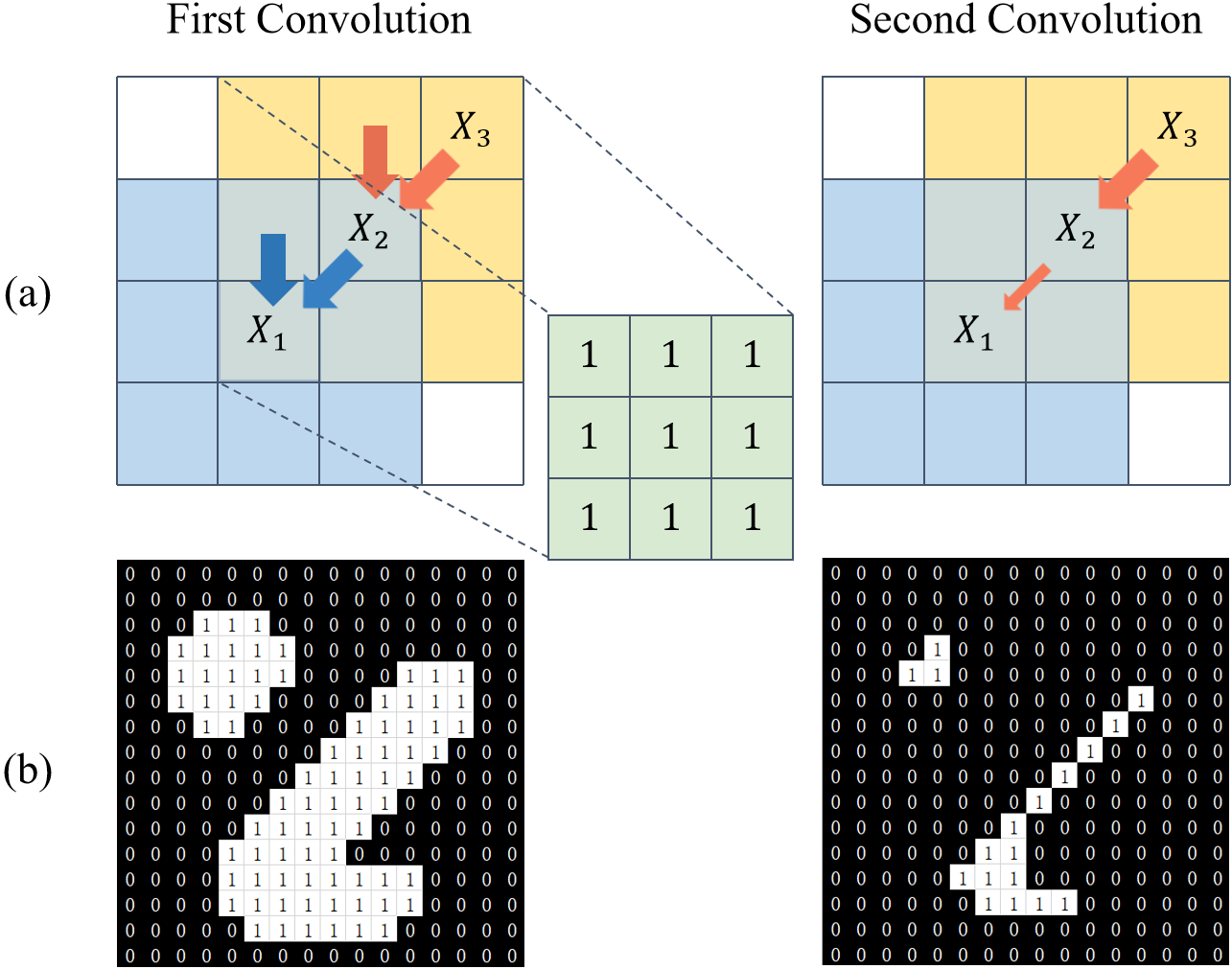}
\end{center}
\caption{Corrosion caused by convolution. When performing the first convolution, the information in the blue area will converge to pixel $X_1$ along the direction of the blue arrow. Similarly, the information in the yellow area will converge to $X_2$. When performing the second convolution, due to the aggregation of $X_3$ information by $X_2$, $X_1$ will also indirectly receive $X_3$ information. The result of corrosion is the shrinkage of the effective area as (b) shown.}
\label{fig:corrosion} 
\end{figure}

On the way to solving the 3D small object tracking, there are two main barriers for current deep learning-based methods: 1) sparse and limited foreground points in the whole scene, which will lead to features of insufficient discrimination; 2) small objects are generally accompanied by narrow or pocket bounding boxes, which induces small objects more sensitive to disturbances. The first issue derives from two aspects. On the one hand, LiDAR emits fan-shaped laser rays and receives intensity signals to generate points after hitting objects. Due to the tiny size of small objects, rays have a lower probability of covering the target, leading to a lower density of foreground points. On the other hand, farthest point sampling is always applied to reduce the number of points during the extraction of high-level features. It uniformly collects points from the whole space, but the foreground points belonging to small objects only concentrate on a small area, which further results in fewer foreground points after sampling. Regarding the second barrier, it will bring the box-sensitivity phenomenon that Success metrics tend to decline significantly even if the predicted box deviates slightly. Essentially, Success metrics are related to the Intersection-Over-Union (IoU) between ground truth and predicted boxes, which is easily affected by the prediction bias. This requires the predicted boxes more precise to keep the Success metric tolerable. We find that an important reason for the inaccurate prediction is attributed to the information erosion caused by convolution. As Fig~\ref{fig:corrosion} shows, convolution allows information to flow between different pixels, which will erode the key characteristics of small objects. Owing to the small object occupying fewer pixels, its features are more easily drowned by invalid features. Based on the above analysis, the direction of improvement to small object tracking can be summarized in the following two aspects: \emph{increasing the density of foreground points}, and \emph{decreasing erosion caused by convolution}.

To obey the first principle, we explore feasible optimization solutions. Recently, ISBNet \cite{Ngo2023ISBNetA3} proposed an instance-wise encoder, which replaces the farthest point sample with an instance-aware sample so that more foreground points could be reserved. Although this sampling strategy could guarantee the density of foreground points, but cannot avoid information loss caused by down-sampling. Taking inspiration from the opposite perspective, we adopt the reverse operation of sampling to increase the density. we propose a target-awareness prototype mining module that adopts the reconstruction mechanism of the masked decoder to mine target prototype information from the whole feature space. This module not only increases the number of foreground points but also restores the individidual geometric information of the target to some extent so that it is capable of enhancing the discrimination of foreground points that will facilitate the subsequent location of small objects. Regarding the second improvement principle, we introduce a regional grid subdivision module to recover fine-grained features of the search region based on the ViT and pixel shuffle layers.
As Tab~\ref{table:voxel_size} shows, shrinking the voxel size of voxelization can improve the Success of small objects. Intuitively, shrinking the voxel size can increase the resolution of the bird's eye view, and higher-resolution images with clearer and richer details are conducive to tracking. Theoretically, more pixels slow down the convolutional information flow and retard corrosion. However, the reduction of voxel size will exponentially increase computing resources. To balance effectiveness and efficiency, one may consider using sparse convolution \cite{Graham20173DSS}. It only performs convolution operations on non-empty pixels and avoids interference with invaluable pixels, but there are invalid pixels rather than non-empty pixels in bird's eye view maps, which will absorb features of interfering substances or unrelated targets. Therefore, we design a regional grid subdivision module that lifts the low-resolution bird’s eye view map to a high-resolution one by a pixel shuffle layer \cite{Shi2016RealTimeSI}. Because the entire process resource consumption is concentrated in the voxelization step, our method can improve accuracy with minimal resource consumption. Finally, we summarize the function of the proposed modules in Fig~\ref{fig:contribution}, which simply demonstrates the advantages of our method.

Considering the experiment on pedestrian category alone is not adequate to verify the robustness of the model against small targets, we elaborately design scaling experiments that scale other categories of objects into similar sizes with pedestrians to explore the adaptability of existing models to small objects. This work devotes to giving an optimization method to improve the tracking performance of small objects. To our knowledge, this paper is the first to bring the concept of small objects into point cloud single object tracking.

In summary, our main contributions are three-fold:
\begin{itemize}
\item We introduce the definition of small object tracking in point cloud scenes and analyze the challenges brought by small objects to 3D single object tracking. To adapt to small objects, the model needs to overcome the low concentration of foreground points and the erosion of features caused by convolution.
\item We propose a target-awareness prototype mining module and a regional grid subdivision module. The former increases the number of foreground points without losing information while the latter decreases erosion without bringing additional computational burden.
\item We design a scaling experiment to compare the robustness of various methods against small objects. Our method has achieved impressive results in both conventional and scaling settings.
\end{itemize}

%% file: sections/relatedworks.tex
\begin{figure*}
\begin{center}
\includegraphics[scale=0.51]{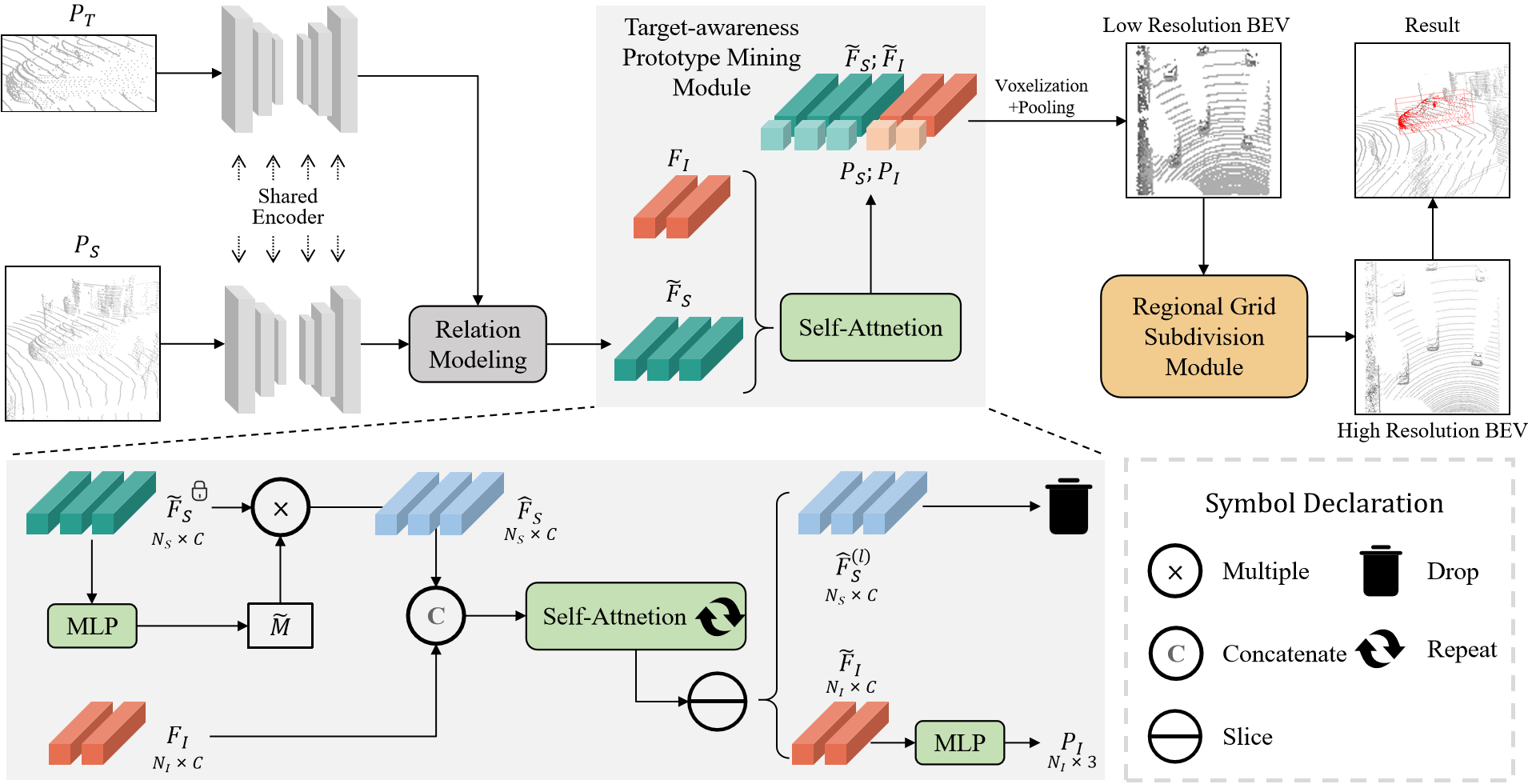}
\end{center}
\caption{The overall architecture of our method. The template point cloud and the search region point cloud are fed into a shared encoder to generate respective geometric features. Then the template features are embedded into search region features by the relation modeling module. Subsequently, the target-awareness prototype mining module will highlight the foreground features and enrich them through several self-attention layers. Finally, the target position is predicted by a bev-based detection head accompanied by a regional grid subdivision module.}
\label{fig:structure} 
\end{figure*}

\section{Related Work}
\label{sec:related}
\subsection{3D Single Object Tracking}
SC3D \cite{Giancola2019LeveragingSC} is the beginning of using deep neural networks for point cloud single object tracking, which calculates the similarity score of candidates with respect to the template, and selects the one with the highest score as the final result. However, its results are subject to the generation of candidates that cannot be integrated into the whole pipeline in an end-to-end manner. Afterward P2B \cite{Qi2020P2BPN} formulates an end-to-end framework, which embeds template features into search features and then feeds the fused features into a VoteNet \cite{Qi2019DeepHV} to generate predicted bounding boxes. P2B has become a milestone whose paradigms are forking by researchers subsequently. MLVSNet \cite{Wang2021MLVSNetMV} adopts Hough voting on multi-level features to adjust targets at different scales. Treating the ground-truth bounding box given in the first frame as a strong cue, BAT \cite{Zheng2021BoxAwareFE} improves the information embedding processing by geometric prior, assisting the model in learning more accurate features. Observing the sparse issue in the voting-based head for predicting the bounding box, V2B \cite{Hui20213DSV} replaces the voting-based detection head with a bev-based one, which projects search points onto a dense bird’s eye view (BEV) feature map to tackle the sparsity of point clouds. With the popularity of transformer \cite{Vaswani2017AttentionIA} in computer vision, LTTR \cite{Cui20213DOT}, PTT \cite{Shan2021PTTPM}, PTTR \cite{Zhou2021PTTRR3}, Trans3DT \cite{Wang2022Accurate3S}, STNet \cite{Hui20223DST} attempt to embed attention module into the original model. Among them, STNet is the most impressive one that
simultaneously leverages the idea of Transformer on both feature extraction and information embedding module. Though achieving considerable performances based on global and local encoder design, it is still struggling with small object tracking. In light of this, we aim to address the problem with the TAPM and RGS modules. 

Recently, to fully take advantage of the background, some methods that consume the whole template instead of a target-only template have been arisen. M2-Track \cite{Zheng2022Beyond3S} and CXTrack \cite{Xu2022CXTrackI3} start using the complete previous frame to enhance the feature fusion process. M2-Track generates coarse bounding boxes by predicting the relative motion between the previous frame and the current frame and then it refines the coarse bounding box to get a precise one in the second stage. CXTrack directly feeds the previous frame and current frame point cloud into their well-designed information embedding module. If first identifies targets by predicting centers of the bounding boxes and segmentation of the foreground, afterwards, fusion features output from the previous module will be utilized to predict results by the transformer-based X-RPN module. Despite the excellent tracking results of these methods, for the sake of fairness, we mainly compare our method with traditional methods that do not introduce additional information.

\subsection{Small Objects Researches}

In the field of visual conception, small objects have always been a challenging problem. We summarize some prevalent and key methods for small objects from three perspectives: multi-scale feature learning, context-based methods, and data augmentation.

\noindent\textbf{Multi-scale Feature Learning} Such methods recon that small object features are prone to be masked by other object features. Therefore, to highlight the small object features, researchers proposed the feature pyramids network \cite{Lin2016FeaturePN}, which aggregates information in different feature layers thereby multi-scale features could produce an effect equally. Although PANet \cite{Liu2018PathAN}, AugFPN \cite{Guo2019AugFPNIM}, TridentNet \cite{Li2019ScaleAwareTN} and many other similar methods are effective, the difference lies in the details of the model rather than the central idea.

\noindent\textbf{Context-based Methods} Due to the inherent characteristics of small objects, it is difficult to obtain sufficient information from the target. Therefore, some methods look forward to assisting tasks with inter-object relationships. ION \cite{Bell2015InsideOutsideND} integrates contextual information outside the region of interest (ROI) using spatial recurrent neural networks. PyramidBox \cite{Tang2018PyramidBoxAC} proposes a novel context-assisted single-shot face detector. Relation Networks \cite{Hu2017RelationNF} designs an object relation module that processes a set of objects simultaneously through interaction between their appearance feature and geometry.

\noindent\textbf{Data Augmentation} This kind of method can be divided into two categories. One is traditional data augmentation, which increases the frequency of tiny targets in the training phase to deal with small objects. In Stitcher \cite{Chen2020StitcherFD}, images are resized into smaller components and then stitched into the same size as regular images. Augmentation \cite{Kisantal2019AugmentationFS} proposed to oversample those images with small objects and augment each of those images by copy-pasting small objects many times. The other is the GAN-based method, which utilizes GAN to generate high-resolution images or high-resolution features. In SOD-MTGAN \cite{Bai2018SODMTGANSO}, the generator is a super-resolution network that can up-sample small blurred images into fine-scale ones and recover detailed information for more accurate detection. Perceptual GAN \cite{Li2017PerceptualGA} improves small object detection by narrowing the representation difference of small objects from the large ones. BFFBB \cite{Noh2019BetterTF} proposes a novel feature-level super-resolution approach that not only correctly addresses these two desiderata but also is integrable with any proposal-based detectors with feature pooling.

\noindent\textbf{Special Design} S3FD \cite{Zhang2017S3FDSS} aims to solve the common problem that anchor-based detectors deteriorate dramatically as the objects become smaller by special care given to small faces in various stages of training. Feedback-driven \cite{Liu2021FeedbackdrivenLF} uses the loss distribution information as the feedback signal guiding the gradient feedback process.

%% file: sections/methods.tex
\section{Methodology}  
\label{sec:method}

\subsection{Problem Defination}
\label{sec:PD}
Given a point cloud of the first frame with the bounding box of an arbitrarily specified object ( template), 3D single object tracking aims to locate the designated target in the following frames. Most prevailing tracking methods comply with siamese-paradigm by which the target will be retrieved with the template cropped from the first frame. The task can be casted as finding a $Tracker$ meeting with Eqn~\ref{eqn:1},
\begin{equation}
\label{eqn:1}
Tracker\left(P_S, P_T\right)=B_S,
\end{equation}
where $P_T$ denotes the set of template points. $P_S$ and $B_S$ denote the set of search region points and the bounding box of the target in the current frame respectively. In addition, the physical size of the target usually remains unchanged in the 3D space thus the bounding box is represented by $B=(x, y, z, \theta)$, which consists of the coordinate of the box center $(x,y,z)$ and the orientation angle $\theta$ around the up-axis. Specifically, the search region points $P_S \in \mathbb{R}^{N_S \times 3}$ and the template points $P_T \in \mathbb{R}^{N_T \times 3}$ will be fed into a shared encoder to acquire individual geometric features $F_S \in \mathbb{R}^{N_S \times C}$ and $F_T \in \mathbb{R}^{N_T \times C}$. Then, the template features will be embedded into search features by a relation modeling module to generate fusion features $\widetilde{F}_S \in \mathbb{R}^{N_S \times C}$. Finally, a detection head will utilize fusion features to predict the final results. Here we focus on the bev-based detection head, which projects the features into bird's eye view maps $V \in \mathbb{R}^{H \times W \times C}$ by voxelization and pooling operation, then use a series of 2D convolution layers to predict Hot Map $\mathcal{H} \in \mathbb{R}^{H \times W \times 1}$, Offset $\mathcal{O} \in \mathbb{R}^{H \times W \times 3}$ and Z-axis $\mathcal{Z} \in \mathbb{R}^{H \times W \times 1}$ separately as described in V2B \cite{Hui20213DSV}.

As discussed in Section~\ref{sec:introduction}, existing methods ignore challenges incurred by small objects. To this end, we propose two effective modules: the target-awareness prototype mining module and the regional grid subdivision module. We will illustrate the implementation of two designed modules in Section~\ref{sec:TAPM} and Section~\ref{sec:RGS} respectively. What's more, details and the loss functions will be given in Section~\ref{sec:loss}.

\subsection{Target-awareness Prototype Mining}
\label{sec:TAPM}
To highlight the presence of the target in the search region, we propose the target-awareness prototype mining (TAPM) module, which is based on the reconstruction mechanism of the masked decoder. Specifically, as Fig~\ref{fig:structure} shown, template points $P_T$ and search region points $P_S$ are firstly fed into shared-encoder to generate their respective features and then they will pass through relation modeling module to obtain fusion features $\widetilde{F}_S$. Subsequently, the fusion features $\widetilde{F}_S$ will be sent into the TAPM module accompanied with learnable substrate features $ F_I  \in \mathbb{R}^{N_I \times C}$. It is worth noting that we cut off the gradient feedback process from $\widetilde{F}_S$ because we intend to regard it as a wizard to guide substrate features $F_I$ transformation to prototype features without affecting the original feature fusion process. Specifically, a Multi-layer Perceptron (MLP) is employed to generate the mask $\widetilde{M} \in \mathbb{R}^{N_S \times 1}$ and then $\widetilde{M}$ multiply as weight into fused features to obtain enhanced fusion features $\widehat{F}_S \in \mathbb{R}^{N_S \times C}$ as Eqn~\ref{eqn:2} shows,
\begin{equation}
\label{eqn:2}
\widetilde{M}=\sigma\left(F W^T+b\right), \widehat{F}_S=\widetilde{M} \cdot \widetilde{F}_S,
\end{equation}
where $\sigma$ is sigmoid function, $\cdot$ is element-wise product. The value of this step is to enhance the features of the foreground points and weaken the features of the background points. Afterward the concatenation of substrate features $F_I$ and fusion features $\widehat{F_S}$ will pass through self-attention layers and iterate $l$ times as Eqn~\ref{eqn:3}
\begin{equation}
\label{eqn:3}
\left[\widehat{F}_S^{(i)}, F_I^{(i)}\right]=\operatorname{Attn}\left(\left[\widehat{F}_S^{(i+1)}, F_I^{(i+1)}\right]\right),
\end{equation}
where $1 \leq i \leq l$, $[,]$ means concatenate operation, $\widehat{F}_S^0 = \widehat{F}_S, F_I^0 = F_I$. 

Leveraging a series of self-attention layers, substrate features repeated interaction with fusion features and finally transfer into prototype features $ \widetilde{F}_I = F_I^(l)$. The teacher features $\widehat{F}_S^{(l)}$ will be dropped after completing its role of guidance. Finally, prototype features $\widetilde{F}_I$ will be used to predict the coordinate of completion points $P_I \in \mathbb{R}^{N_I \times 3}$ by a MLP layer and the enhanced point cloud $[P_S, P_I; \widetilde{F}_S, \widetilde{F}_I]$ will be sent to downstream task. To ensure that the prototype points fall on the target, we introduce Chamfer Distance(CD) loss constraint coordinates, as detailed in Section~\ref{sec:loss}.

\begin{figure}
\begin{center}
\includegraphics[scale=0.32]{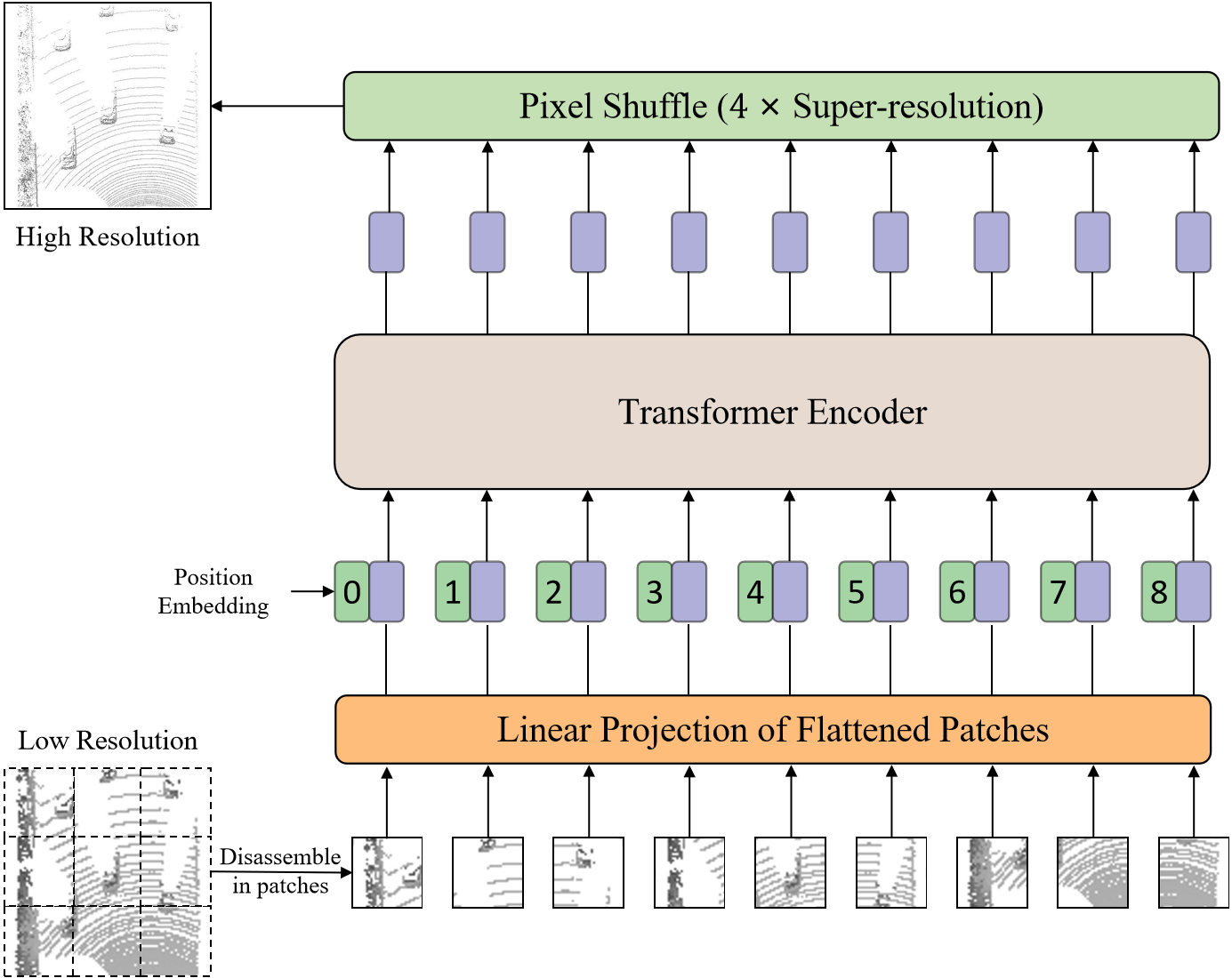}
\end{center}
\caption{The architecture of the regional grid subdivision module, which consists of a Vit layer and a pixel shuffle layer. Firstly, the bird's eye view feature map decouples into patches, and the patches will be flattened and added with position embedding. They will interact and generate reconstructed features by a series of transformer encoders. Finally, each 128-channel pixel feature is divided into four 32-channel pixel features by the pixel shuffle layer.}
\label{fig:vit}
\end{figure}

\begin{figure*}
\begin{center}
\includegraphics[width=\linewidth]{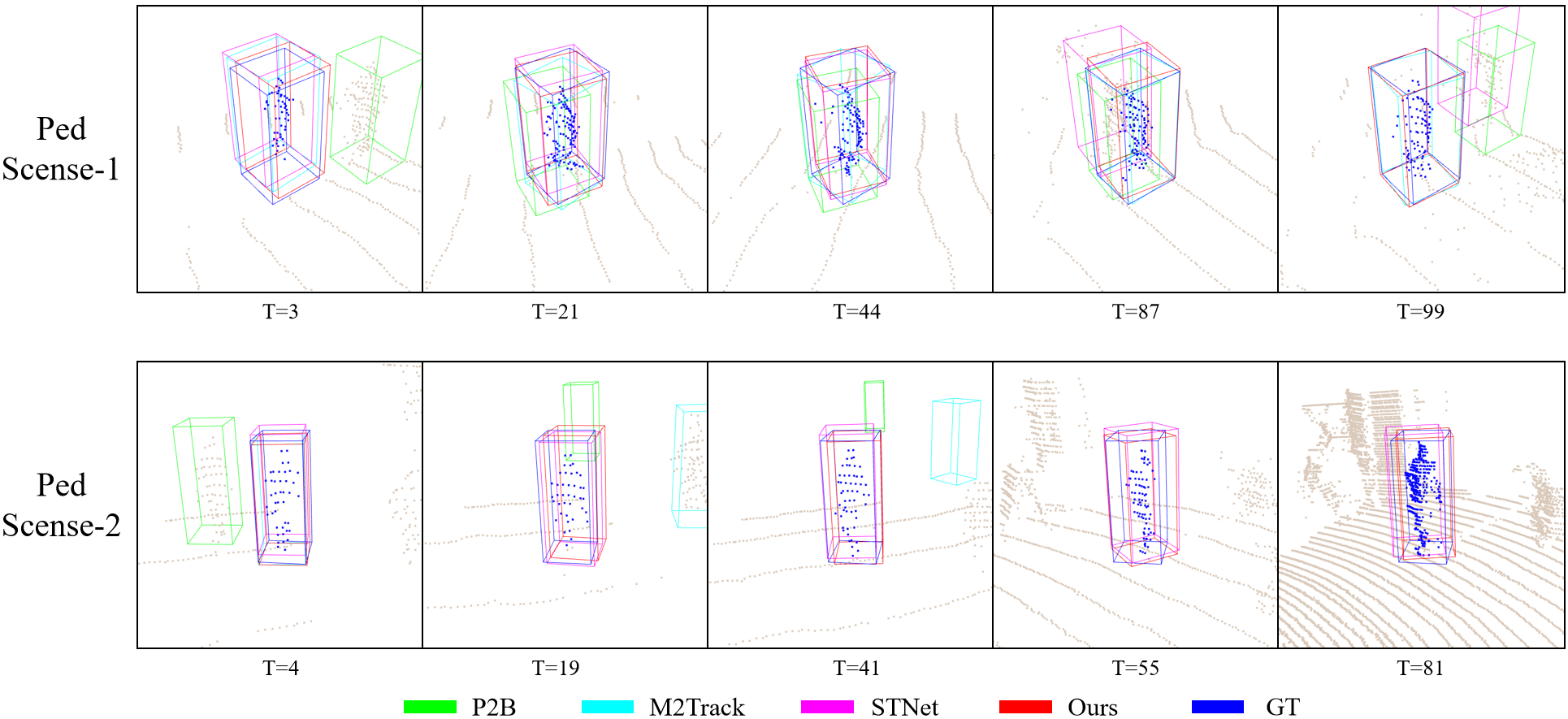}
\end{center}
\caption{Visualization results of the pedestrian category. We selected two typical sequences in the KITTI datasets to compare the tracking results of different methods.}
\label{fig:vis} 
\end{figure*}

\subsection{Regional Grid Subdivision}
\label{sec:RGS}
Small objects are sensitive to prediction bias, so they require more precise positioning than general objects to obtain outsranding results. One of the important reasons for positioning deviation is the information dilution brought by convolution. It has been proven in the field of traditional vision that the higher the resolution, the stronger the robustness of the image to erosion. Inspired by this, we insert a regional grid subdivision (RGS) module into the original bev-based detection head. 

As Fig~\ref{fig:structure} shown, the RGS module is composed of two subparts: a Vit layer \cite{Dosovitskiy2020AnII} and a pixel shuffle layer \cite{Shi2016RealTimeSI}. At first, the enhanced point cloud $[P_S, P_I; \widetilde{F}_S, \widetilde{F}_I]$ is projected into a bird's eye view feature map $V \in \mathbb{R}^{H \times W \times C}$ by voxelization and pooling. Then this map will be patched based on pixels and sent to a Vit layer, whose structure is shown in Fig~\ref{fig:vit}. A pixel shuffle layer is employed to generate a high-resolution BEV feature map $\widetilde{V} \in \mathbb{R}^{2 H \times 2 W \times \frac{C}{4}}$. Finally, $\widetilde{V}$ passes through a series of 2D convolutional layers to generate the final result which consists of Hot Map $\mathcal{H} \in \mathbb{R}^{H \times W \times 1}$, Offset $\mathcal{O} \in \mathbb{R}^{H \times W \times 3}$ and Z-axis $\mathcal{Z} \in \mathbb{R}^{H \times W \times 1}$. The introduction of the Vit layer has two aspects of significance. On the one hand, it reorganizes pixel features into channels so that makes it easier to separate detailed features from the overall during conducting sub-pixel. On the other hand, the Vit layer helps pixel features capture global information and compensates for the lack of global information caused by increased resolution.

\subsection{Loss Functions}
\label{sec:loss}
Following STNet \cite{Hui20223DST}, we employ focal loss $\mathcal{L}_{hm}$, smooth-L1 loss $\mathcal{L}_{off}$, $\mathcal{L}_z$ to constraint $\mathcal{H}$, $\mathcal{O}$, $\mathcal{Z}$ in first stage respectively. At first, some variables need to be declared. The real coordinate of the target center is $(x, y, z)$. $\theta$ indicates the ground truth rotation angle. $v$ is the voxel size and $\left(x_{m i n}, y_{m i n} \right)$ indicates the minimum coordinate of search regions. $c= \left( c_x, c_y \right)$ is 2D target center in x-y plane, where $c_x= \frac{x - x_{m i n}}{2v}$ and $c_y=\frac{y - y_{m i n}}{2v}$. The discrete 2D center $\tilde{c} = (\tilde{c}_x, \tilde{c}_y)$ is defined by $\tilde{c}_x=\left\lfloor c_x\right\rfloor$ and $\tilde{c}_y=\left\lfloor c_y\right\rfloor$, where $\left\lfloor.\right\rfloor$ means rounding down. Then focal loss $\mathcal{L}_{hm}$ between $\mathcal{H}$ and the ground truth $\mathcal{H}^{gt}$, the smooth-L1 loss $\mathcal{L}_{off}$ between $\mathcal{O}$ and the ground truth $\mathcal{O}^{gt}$, the smooth-L1 loss $\mathcal{L}_{off}$ between $\mathcal{Z}$ and the ground truth $\mathcal{Z}^{gt}$ can be calculated, where $\mathcal{H}_{ij}^{gt}=1$ if $\left( i,j \right) =\tilde{c}$ otherwise $\frac{1}{1+\|(i, j)-\widetilde{c}\|}$, $\mathcal{O}_{ij}^{gt}=[ \left( i,j \right)-c,\theta]$ and $\mathcal{Z}_{ij}^{gt} = z$. In addition, to ensure that the prototype points are target-aware, we use Chamfer distance(CD) loss $\mathcal{L}_{cd}$ to constrain interpolation points $P_I$ as Eqn~\ref{eqn:4}

\begin{equation}
\begin{aligned}
\label{eqn:4}
\mathcal{L}_{cd} = \sum_{p_T^i \in P_T} \min _{p_I^j \in P_I}\left\|p_T^i-p_I^j\right\|_2^2  +\sum_{p_I^i \in P_I} \min _{p_T^j \in P_T}\left\|p_I^i-p_T^j\right\|_2^2.
\end{aligned}
\end{equation}
Significantly, $P_T$ is a template point cloud aligned with the current frame target through translation and rotation. Finally, We aggregate all the mentioned losses as our final loss
\begin{equation}
\begin{aligned}
\label{eqn:8}
\mathcal{L} = \lambda_1 (\mathcal{L}_{h m}+\mathcal{L}_{off}) + \lambda_2 \mathcal{L}_z + \lambda_3 \mathcal{L}_{cd},
\end{aligned}
\end{equation}
where $\lambda_1=1$, $\lambda_2=2$, $\lambda_3=1\times 10^{-6}$ for non-rigid objects and $\lambda_3=2 \times 10^{-7}$ for rigid objects.

%% file: sections/experiments.tex
\begin{table*}[htb]
\caption{Comparison with mainstream methods under our scaling setting. We simulate small object tracking by scaling other categories of objects to the same size as pedestrians. Specifically, we scaled the car and van categories by a factor of 0.25 and the cyclist category by a factor of 0.5. In this way, the pixels occupied by all categories of objects in the bird's eye view meet the definition of small objects. "ORIG" refers to the experiment under the original setting. "SC" refers to the experiment under the scaling setting. "GAP" refers to the score gap before and after the change of the experimental setting.}
\centering
\resizebox{\linewidth}{!}{
\begin{tabular}{c|c|cc|cc|cc|ccc}
\hline
                          & Pedestrian                  & \multicolumn{2}{c|}{Car}                & \multicolumn{2}{c|}{Van}                & \multicolumn{2}{c|}{Cyclist}            & \multicolumn{3}{c}{Mean}                                                 \\ \cline{2-11} 
\multirow{-2}{*}{Methods} & ORIG & ORIG & SC        & ORIG & SC        & ORIG & SC        & ORIG & SC        & GAP                            \\ \hline
P2B                       & 28.7/49.6                   & 56.2/72.8                   & 15.4/13.3 & 40.8/48.4                   & 12.2/4.7  & 32.1/44.7                   & 34.9/53.5 & 42.4/60.0                   & 21.3/29.1 & -21.1/-30.9                    \\
BAT                       & 42.1/70.1                   & 60.5/77.7                   & 16.2/20.9 & 52.4/67.0                   & 10.2/9.0 & 33.7/45.4                   & 17.9/29.7 & 51.2/72.8                   & 26.9/41.3 & -24.3/-31.5                      \\
M2Track                   & 61.5/88.2                   & 65.5/80.8                   & 22.4/28.0 & 53.8/70.7                   & 8.6/5.9   & 73.2/93.5                   & 69.7/87.8 & 62.9/83.4                   & 38.9/53.4 & -20.5/-30.7                      \\
STNet                     & 49.9/77.2                   & 72.1/84.0                   & 60.5/82.2 & 58.0/70.6                   & 48.1/77.8 & 73.5/93.7                   & 69.4/96.5 & 61.3/80.1                   & 55.0/79.9 & -6.3/-0.2                      \\ \hline
Ours                      & 58.5/83.4                   & 71.5/84.0                   & 63.5/84.6 & 60.3/74.9                   & 51.1/83.4 & 73.0/93.9                   & 73.8/97.1 & 64.9/83.0                   & 60.4/84.2 & -4.5/1.2                       \\
Improvement               & ↑8.6/↑6.2                   & ↓0.6/-0.0                      & ↑3.0/↑2.4 & ↑2.3/↑4.3                   & ↑3.0/↑5.6 & ↓0.5/↑0.2                   & ↑4.3/↑0.6 & ↑3.6/↑2.9                   & ↑5.4/↑4.3 & ↑1.8/↑1.4 \\ \hline
\end{tabular}}
\label{table:scale_experiment}
\end{table*}

\section{Experiments}
\label{sec:experiments}

\subsection{Experimental Settings}  
\label{sec:implementation}
\noindent\textbf{Dataset.} We conducted extensive experiments on the KITTI \cite{Geiger2012AreWR} and nuScense \cite{Caesar2019nuScenesAM} datasets. KITTI contains 21 training video sequences and 29 test sequences. We follow SC3D \cite{Giancola2019LeveragingSC} and split the training sequences into three parts, 0-16 for training, 17-18 for validation, and 19-20 for testing. For nuScense, we use its validation split to evaluate our model, which contains 150 scenes. In addition, to compare the robustness of existing methods to small objects, we have also proposed a new scaling experimental setup on KITTI. All experiments are conducted with TITAN GPUs. 

\noindent\textbf{Implementation Details.} For the 3D single object tracking task, we enlarge the ground truth bounding box in the current frame by 2 meters to obtain the sub-region and sample 1024 points from it as $P_S$. Simultaneously, we crop and aggregate the target point clouds of the first and previous frames, and then sample 512 points from aggregated points as templates $P_T$. For the shared encoder and relation modeling module, we use the same settings as STNet \cite{Hui20223DST}. For the target-awareness prototype mining module, we set the prototype point number equal to 64 and iterate the self-attention 5 times. For the detection head, we set the voxel size equal to 0.2m and perform 4x super-resolution on bird's eye view feature maps.

\noindent\textbf{Evaluation Metrics.} We follow one pass evaluation \cite{Kristan2015ANP} to evaluate all the methods, which consists of two evaluate metrics: Success and Precision. The former denotes the Area Under Curve for the plot showing the ratio of frames where the Intersection Over Union (IOU) between the predicted and ground-truth bounding boxes is larger than a threshold, ranging from 0 to 1. The latter is defined as the AUC of the plot showing the ratio of frames where the distance between predicted and ground-truth bounding box centers is within a threshold, from 0 to 2 meters.

\noindent\textbf{Scaling experiment.} Given a scene point cloud, we extract the foreground points based on the bounding box. We mark the set of foreground points as $P_f$ and the background as $P_b$. The set of foreground points can be represented by Eqn~\ref{eqn:6}
\begin{equation}
\label{eqn:6}
P_f=\left\{p_i=\left(x+\Delta x_i, y+\Delta y_i, z+\Delta z_i\right)\right\}_{i=1}^{N_f}
\end{equation}
where $(x,y,z)$ is the coordinates of the center point of the bounding box. After scaling, the new foreground point set will be defined as Eqn~\ref{eqn:7} and the set of background points $P_b$ remaining
\begin{equation}
\label{eqn:7}
\widetilde{P}_f=\left\{p_i=\left(x+ r \Delta x_i, y+ r \Delta y_i, z+ r \Delta z_i\right)\right\}_{i=1}^{N_f}
\end{equation}
where $r \in (0,1)$ is scaling rate parameter.

\begin{table}[htb]
\caption{Comparison with state of the arts on Kitti datasets. The results in the cells represent the "Success/Precision" of the corresponding method (row) under the corresponding category (column). The {\color[HTML]{FF0000} red} has the highest Success, the {\color[HTML]{2A9D8F} mint} has the second.}
\centering
\resizebox{\linewidth}{!}{
\begin{tabular}{c|c|c|c|c|c}
\hline
Methods                         & \begin{tabular}[c]{@{}c@{}}Car\\ (6424)\end{tabular} & \begin{tabular}[c]{@{}c@{}}Pedestrian\\ (6088)\end{tabular} & \begin{tabular}[c]{@{}c@{}}Van\\ (1248)\end{tabular} & \begin{tabular}[c]{@{}c@{}}Cyclist\\ (308)\end{tabular} & Mean                             \\ \hline
SC3D                            & 41.3/57.9                                            & 18.2/37.8                                                   & 40.4/47.0                                            & 41.5/70.4                                               & 31.2/48.5                        \\
P2B                             & 56.2/72.8                                            & 28.7/49.6                                                   & 40.8/48.4                                            & 32.1/44.7                                               & 42.4/60.0                        \\
3DSiamRPN                       & 58.2/76.2                                            & 35.2/56.2                                                   & 45.7/52.9                                            & 36.2/49.0                                               & 46.7/64.9                        \\
LTTR                            & 65.0/77.1                                            & 33.2/56.8                                                   & 35.8/45.6                                            & 66.2/89.9                                               & 48.7/65.8                        \\
MLVSNet                         & 56.0/74.0                                            & 34.1/61.1                                                   & 52.0/61.4                                            & 34.3/44.5                                               & 45.7/66.7                        \\
BAT                             & 60.5/77.7                                            & 42.1/70.1                                                   & 52.4/67.0                                            & 33.7/45.4                                               & 51.2/72.8                        \\
PTT                             & 67.8/81.8                                            & 44.9/72.0                                                   & 43.6/52.5                                            & 37.2/47.3                                               & 55.1/74.2                        \\
V2B                             & 70.5/81.3                                            & 48.3/73.5                                                   & 50.1/58.0                                            & 40.8/49.7                                               & 58.4/75.2                        \\
PTTR                            & 65.2/77.4                                            & 50.9/81.6                                                   & 52.5/61.8                                            & 65.1/90.5                                               & 57.9/78.1                        \\
 STNet & {\color[HTML]{2A9D8F} 72.1/84.0} & 49.9/77.2 & 58.0/70.6 & {\color[HTML]{FF0000} 73.5/93.7} & 61.3/80.1 \\
M2Track & 65.5/80.8 & {\color[HTML]{FF0000} 61.5/88.2} & 53.8/70.7 & {\color[HTML]{2A9D8F} 73.2/93.5} & {\color[HTML]{2A9D8F}62.9/83.4} \\ 
Trans3DT & {\color[HTML]{FF0000} 73.3/84.7} & 53.5/79.8 & {\color[HTML]{2A9D8F} 59.2/70.5} & 46.3/56.5 & 62.9/80.7 \\ \hline
Ours & 71.5/84.0 & {\color[HTML]{2A9D8F} 58.5/83.4} & {\color[HTML]{FF0000} 60.3/74.9} & 73.0/93.9 & {\color[HTML]{FF0000} 64.9/83.0} \\ \hline
\end{tabular}}
\label{table:kitti}
\end{table}

\begin{table}[htb]
\caption{Comparison with state of the arts on nuScense datasets. The results in the cells represent the "Success/Precision" of the corresponding method (row) under the corresponding category (column). * means the results of migrating the model to the same environment as ours tested it.}
\centering
\resizebox{\linewidth}{!}{
\begin{tabular}{c|c|c|c|c|c}
\hline
Methods  & \begin{tabular}[c]{@{}c@{}}Car\\ (15578)\end{tabular} & \begin{tabular}[c]{@{}c@{}}Pedestrian\\ (8019)\end{tabular} & \begin{tabular}[c]{@{}c@{}}Truck\\ (3710)\end{tabular} & \begin{tabular}[c]{@{}c@{}}Bicycle\\ (501)\end{tabular} & Mean      \\ \hline
SC3D     & 25.0/27.1                                             & 14.2/17.2                                                   & 25.7/21.9                                              & 17.0/18.2                                               & 21.8/23.1 \\
P2B      & 27.0/29.2                                             & 15.9/22.0                                                   & 21.5/16.2                                              & 20.0/26.4                                               & 22.9/25.3 \\
BAT      & 22.5/24.1                                             & 17.3/24.5                                                   & 19.3/15.8                                              & 17.0/18.8                                               & 20.5/23.0 \\
V2B      & 31.3/35.1                                             & 17.3/23.4                                                   & 21.7/16.7                                              & 22.2/19.1                                               & 25.8/29.0 \\
STNet    & 32.2/36.1                                             & 19.1/27.2                                                   & 22.3/16.8                                              & 21.2/29.2                                               & 26.9/30.8 \\
Trans3DT & 31.8/35.4                                             & 17.4/23.3                                                   & 22.7/17.1                                              & 18.5/23.9                                               & 26.2/29.3 \\
P2B*     & 24.1/24.6                                             & 16.5/20.0                                                   & 18.8/13.1                                              & 17.5/18.9                                               & 21.1/21.6 \\
M2Track* & 27.2/28.3                                             & 16.4/18.9                                                   & 20.1/16.5                                              & 16.9/16.6                                               & 23.0/23.8 \\
STNet*   & 25.5/27.0                                             & 14.9/16.3                                                   & 18.9/13.3                                              & 17.0/16.4                                               & 21.4/21.9 \\ \hline
Ours     & 25.6/27.5                                             & 15.3/17.4                                                   & 12.7/18.5                                              & 17.5/18.4                                               & 20.8/23.2 \\ \hline
\end{tabular}}
\label{table:nuscense}
\end{table}

\subsection{Results}
\label{sec:results}
\noindent\textbf{Experiments under scaling settings.} We retrain P2B \cite{Qi2020P2BPN}, BAT \cite{Zheng2021BoxAwareFE}, V2B \cite{Hui20213DSV}, STNet \cite{Hui20223DST}, M2Track \cite{Zheng2022Beyond3S} under scaling setting and the results are shown in the Tab~\ref{table:scale_experiment}. After scaling, the Success of all methods decrease significantly, which is consistent with the box-sensitivity phenomenon of small objects. By joint analysis on this table, it is not difficult to find more performance degradation in point-based methods (P2B, BAT, M2Track). The reason is that such methods require selecting candidates through farthest point sampling, however, scaling shrinks the covering volumn of the target and makes it more difficult to obtain points which fall on the targets. Compared to point-based methods, bev-based methods perform more stably, as they do not require sampling during the prediction phase. Our method is 3.6 Success higher than the benchmark (STNet) under the original setting, and the gap widens to 5.4 when migrating to the scaling setting. This fact reflects that our method is more robust than the benchmark.

\noindent\textbf{Experiments on the KITTI dataset.} Except for scaling experiments, our method has also achieved commendable results under conventional settings. We comprehensive comparisons on the KITTI dataset with previous state-of-the-art methods, including SC3D \cite{Giancola2019LeveragingSC}, P2B \cite{Qi2020P2BPN}, 3DSiamRPN \cite{Fang20213DSiamRPNAE}, LTTR \cite{Cui20213DOT}, MLVSNet \cite{Wang2021MLVSNetMV}, BAT \cite{Zheng2021BoxAwareFE}, PTT \cite{Shan2021PTTPM}, V2B \cite{Hui20213DSV}, PTTR \cite{Zhou2021PTTRR3}, STNet \cite{Hui20223DST}, M2-Track \cite{Zheng2022Beyond3S} and Trans3DT \cite{Wang2022Accurate3S}. 
As illustrated in Tab~\ref{table:kitti}, the average Success of our methods exceeds all the above. Compared with the benchmark (STNet), our method has significantly improved performance in the pedestrian category, which owns to specialized optimization of small objects. Unlike M2Track, which belongs to motion-based methods, our method is still based on geometric shape matching. Therefore, albeit our method is slightly inferior to M2Track in the pedestrian category, its performance on rigid objects with clear geometric structures is significantly better than M2Track. 

\noindent\textbf{Experiments on the nuScense dataset.} To compare the generalization ability, we test the KITTI pre-trained model on the nuScenes dataset. Unfortunately, due to the sensitivity of the target-awareness prototype mining module to changes in distribution, our method did not achieve state-of-the-art results as Tab~\ref{table:nuscense} shown.

\begin{figure}
\begin{center}
\includegraphics[scale=0.35]{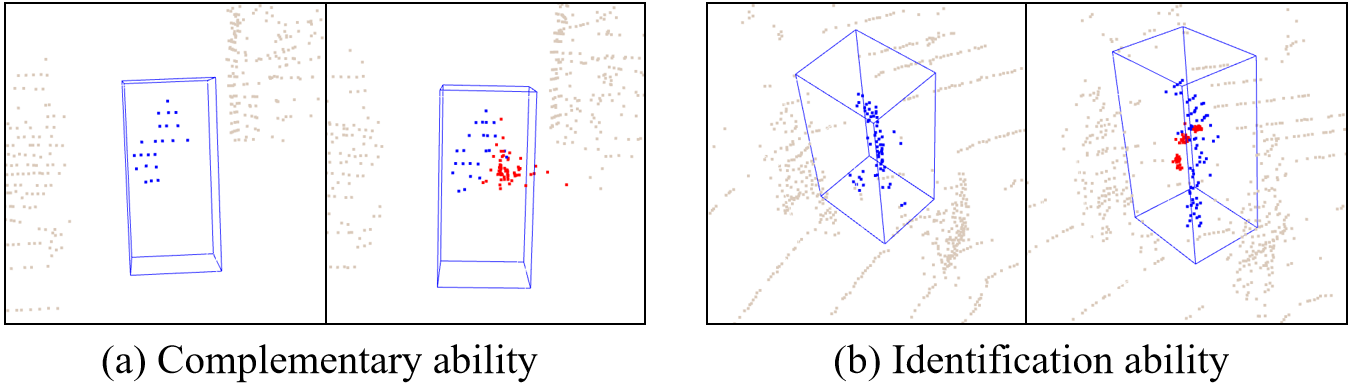}
\end{center}
\caption{Visualization of interpolation points effects. The interpolation points generated by TAPM can restore the shape of the target. More importantly, it can correctly identify the target from intra-class distractors.}
\label{fig:interpolation}
\end{figure}

\begin{figure*}
\begin{center}
\includegraphics[width=\linewidth]{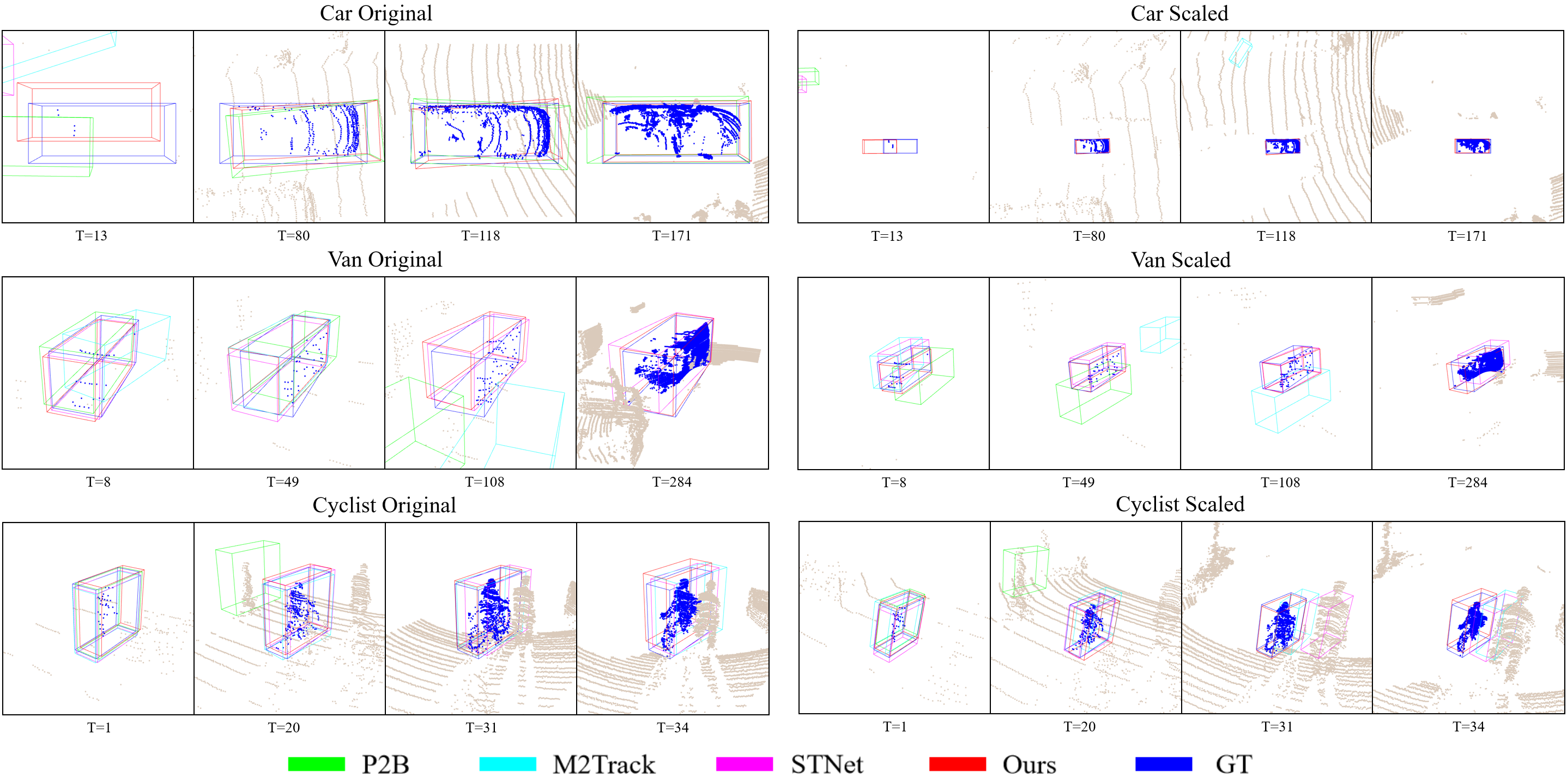}
\end{center}
\caption{Visualization results of scaling experiments. The same scenes were selected before and after scaling, so comparing the two groups of images on the left and right can observe the difference in the effect before and after scaling.}
\label{fig:vis_sc} 
\end{figure*}

\noindent\textbf{Visualization.} Finally, we compare the performance between different methods qualitatively. Comparing the two sets of pictures in Fig~\ref{fig:vis}, when there is no intra-class distractor, most methods can locate the target accurately. However, when there are intra-class distractors, P2B and STNet are misled by the bystander in "Ped Scense-1", while P2B and M2Track are misled in "Pedestrian Scene-2". In addition, although both STNet and ours locate targets successfully in "Pedestrian Scene-2", our bounding boxes are more closely aligned with the ground truth. Comparing the results between the two experimental settings in Fig~\ref{fig:vis_sc}, we can see that the proposed approach has two major advantages. One is that our method can effectively deal with sparsity. For the first scene (Car) in the first row, the foreground points are extremely sparse in the 13th frame (T=13), but in such an extreme case, our method still successfully seeks the target out while other methods fail. The other, which is also the main contribution of this paper, is that our approach is more robust for small objects than other trackers. For "Car Scaled", other methods lose the target in the first few frames, which makes it impossible to track the target even if the point cloud becomes dense in subsequent frames. For "Van Scaled", STNet and our method maintain superior while point-based methods are quite confused about where the target is. For "Cyclist", all methods have achieved remarkable results except P2B under the original setting while STNet and M2Track obtain failure cases caused by the disturbance under the scaling setting. We believe that the robustness of our method to intra-class distractors comes from the prototype mining ability of the TAPM module to complete and recognize the target. As Fig~\ref{fig:interpolation} shows, the TAPM module reconstructs the pedestrian's right arm in Fig(a), and it accurately inserts the points around the target without being misled by other pedestrians in Fig(b).

\begin{figure}
\begin{center}
\includegraphics[scale=0.6]{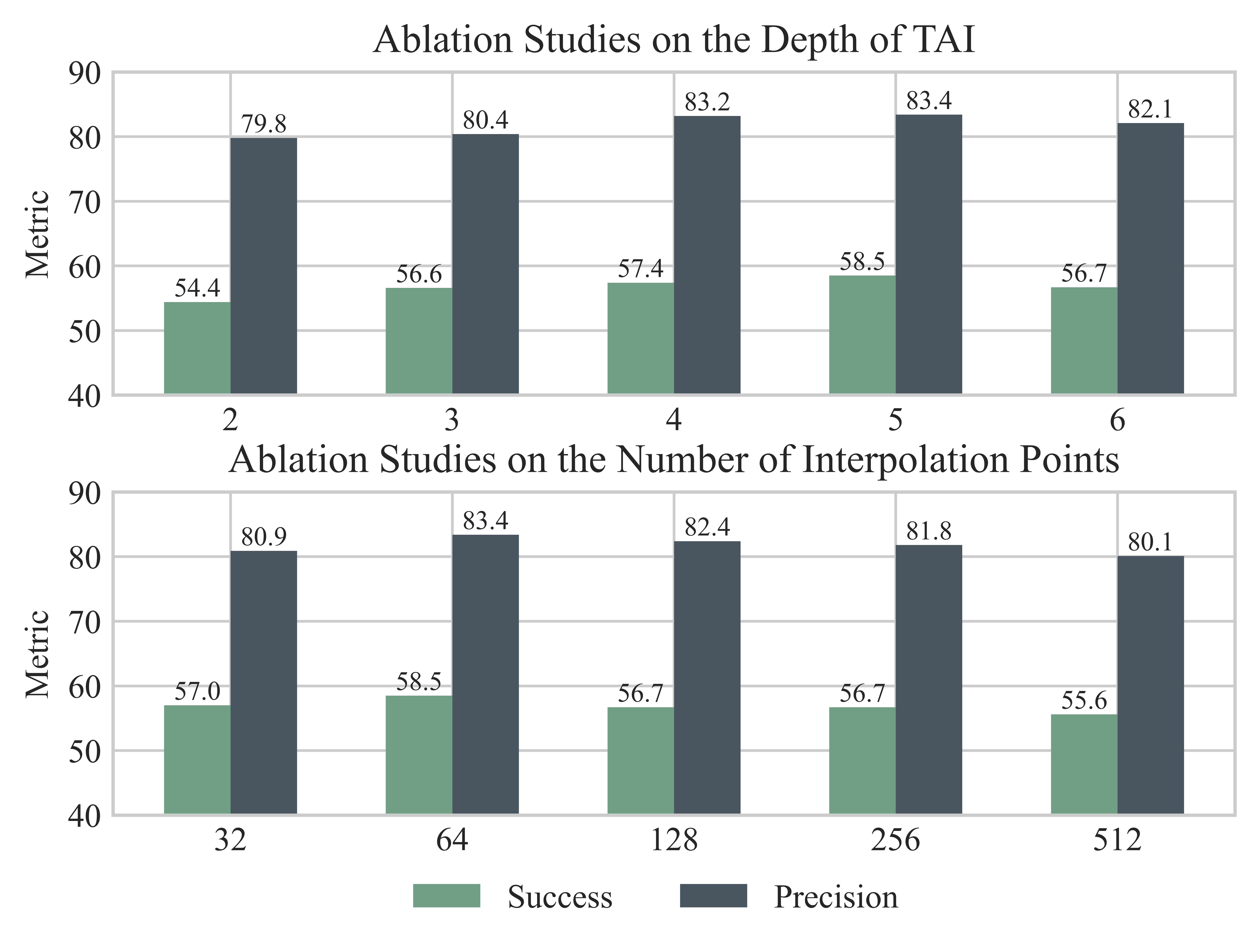}
\end{center}
\caption{Exploration of the TAPM module. Two slides show the effect of the depth of self-attention and the number of interpolation points on the model respectively.}
\label{fig:layer} 
\end{figure}

\begin{table}[htb]
\caption{Ablation studies of different voxel sizes. "STNet-0.2" indicates that the voxel size was set to 0.2m for the experiment with other settings unchanged.}
\centering
\resizebox{\linewidth}{!}{
\begin{tabular}{l|c|c|c|c|c}
\hline
\multicolumn{1}{c|}{Methods} & \begin{tabular}[c]{@{}c@{}}Car\\ (6424)\end{tabular} & \begin{tabular}[c]{@{}c@{}}Pedestrian\\ (6088)\end{tabular} & \begin{tabular}[c]{@{}c@{}}Van\\ (1248)\end{tabular} & \begin{tabular}[c]{@{}c@{}}Cyclist\\ (308)\end{tabular} & Mean      \\ \hline
STNet-0.2                    & 70.8/82.6                                            & 55.4/79.9                                                   & 39.0/45.7                                            & 71.6/93.9                                               & 61.3/78.4 \\
STNet-0.3                    & 70.5/82.7                                            & 51.4/78.8                                                   & 56.5/67.0                                            & 72.9/94.0                                               & 61.0/79.9 \\
STNet-0.4                    & 69.2/82.0                                            & 46.0/74.0                                                   & 58.0/68.1                                            & 73.3/94.2                                               & 58.3/77.6 \\ \hline
Ours-0.2                     & 71.5/84.0                                            & 58.5/83.4                                                   & 60.3/73.9                                            & 73.0/93.9                                               & 64.9/83.0 \\
Ours-0.3                     & 71.4/84.2                                            & 54.5/81.1                                                   & 60.0/73.2                                            & 73.2/94.1                                               & 63.1/82.1 \\
Ours-0.4                     & 69.0/81.7                                            & 50.7/78.1                                                   & 59.7/72.7                                            & 74.0/94.5                                               & 60.3/79.6 \\ \hline
\end{tabular}}
\label{table:voxel_size}
\end{table}

\begin{table}[htb]
\caption{The effectiveness of the proposed module. \checkmark and \textbackslash{} refer to w/o corresponding  module respectively. The values show the results test on the Pedestrian category.}
\centering
\resizebox{\linewidth}{!}{
\begin{tabular}{c|c|c|c|c}
\hline
\multicolumn{1}{l|}{TAPM Module} & \multicolumn{1}{l|}{PixelShuffle} & \multicolumn{1}{l|}{ViT Layer} & Succuss                        & Precision                      \\ \hline
\textbackslash{}                & \textbackslash{}                   & \textbackslash{}               & 51.6                        & 73.6                        \\
\checkmark                               & \textbackslash{}                   & \textbackslash{}               & 51.7                           & 74.9                           \\
\textbackslash{}                & \checkmark                                 & \textbackslash{}               & 49.8 & 70.7 \\
\textbackslash{}                & \textbackslash{}                   & \checkmark                              & 56.0                           & 80.7                           \\
\checkmark                               & \checkmark                                  & \textbackslash{}               & 54.2                           & 75.9                           \\
\textbackslash{}                & \checkmark                                  & \checkmark                              & 56.4                           & 82.4                           \\
\checkmark                               & \textbackslash{}                   & \checkmark                              & 56.4                        & 81.5                        \\
\checkmark                               & \checkmark                                  & \checkmark                              & 58.5                           & 83.4                           \\ \hline
\end{tabular}}
\label{table:module_effect}
\end{table}

\subsection{Ablation Studies}
\label{sec:ablation}
To explore the factors that affect the model results, we conducted ablation experiments on the KITTI dataset.

\noindent\textbf{Voxel size.} Voxel size is the side length of the cube space occupied by each voxel during voxelization. Increasing the value of voxel size will encode the space into a higher resolution volume, and it will increase the resolution of the bird's eye view map which is obtained by pooling the volume. As discussed in Section~\ref{sec:introduction}, decreasing the voxel size can relieve convolutional corrosion and improve accuracy. Tab~\ref{table:voxel_size} shows the gap in average tracking accuracy between STNet and our method under different voxel sizes. When the voxel size of STNet decreases from 0.4m to 0.2m, the mean Success increased from 58.3 to 61.3, among the most obvious benefits are the Pedestrian category. However, as the resolution increased, the performance of STNet in the Van category degraded. This may be because smaller voxel units do not efficiently encode the detailed features of larger objects. In contrast, our method solves this problem by adding the vit layer to make the network more powerful in encoding pixel features. In addition to the comparison of effects, the video memory consumed by the training model cannot be ignored. Although STNet-0.2 can train with 84 batches (ours-0.2 is 42) at a time, the accuracy is much lower than ours, and the computation amount increases exponentially when the voxel size reduces to 0.1. STNet-0.1 can only train with 27 batch size which makes it very difficult for STNet to achieve our accuracy by continuing to reduce the voxel size.

\noindent\textbf{Model components.} We investigate the effectiveness of the proposed module with Tab~\ref{table:module_effect} on the Pedestrian category. It is noteworthy that adding only a pixel shuffle layer will lead to side effects. That is because pixel shuffle splits one 128-channel pixel feature into four 32-channel pixel features so that the structure of each pixel corresponding location cannot be efficiently encoded. After adding the Vit layer which brings more powerful coding capability, the pixel shuffle layer will be a boost to the whole network.

\noindent\textbf{TAPM module.} In this part, we explore the influence of TAPM module hyperparameters on the results. As Fig~\ref{fig:layer} shows, as the self-attention depth of the TAPM module increases from 0 to 5, the metric increases from 54.4/79.8 to 58.5/83.4. After the depth reaches 5, if continuously increasing, the metric will decline. Similarly, the more number of interpolation points does not always mean better performance. When the number of interpolation points is set to around 64, the model can maintain the best performance.

%% file: sections/limitations.tex
\section{Limitations}
There are mainly two short boards of our model. The first is that the ability of the TAPM module to grasp shape is limited. Although the interpolation points can restore the shape of the object to some extent, they cannot restore the details. On the one hand, it is limited by the quality of the template (the template obtained by cropping is generally incomplete), and on the other hand, the effectiveness of using only CD distance constraints on interpolation points is limited. The second is that the generalization of the model to the dataset is sacrificed to perceive the target. Our model trained on 32-line Laser Radar datasets (KITTI) can not adapt to 64-line Laser Radar datasets (nuScense). 

%% file: sections/conclusion.tex
\section{Conclusion}
In this work, we define small objects in 3D single object tracking and provided two modules specifically designed to improve the tracking performance of small objects: the target-awareness prototype mining module and the regional grid subdivision module. The target-awareness prototype mining module extracts the complete target prototype from the whole search region and reconstructs the target by interpolating points. The regional grid subdivision module refines low-resolution bird's eye view maps into high-resolution one and decreases erosion, achieving accurate positioning with pixel shuffle and Vit layers. Finally, abundant experiments demonstrate that our method can effectively improve the tracking accuracy of small objects while ensuring that it does not affect the other types of targets.